\documentclass[twoside,11pt]{article}
\usepackage{journal2e_v1}

\usepackage{amsmath}
\usepackage{amssymb}
\usepackage{mathrsfs}

\usepackage{epsf}
\usepackage{epsfig}
\usepackage{subfigure}
\usepackage{algorithm}
\usepackage{multirow}
\usepackage{multicol}

\usepackage{graphicx} 

\usepackage{algorithm}
\usepackage{algorithmic}

\usepackage[framemethod=tikz,innertopmargin=7pt]{mdframed}

\usepackage[pagebackref=true,breaklinks=true,letterpaper=true,colorlinks,bookmarks=yes]{hyperref}

\newtheorem{alg}{Algorithm}

\newcommand{\uvec}{\textup{uvec}}
\newcommand{\sym}{\textup{sym}}
\newcommand{\diag}{\text{diag}}

\def\E{{\bf E}}
\def\e{{\epsilon}}

\renewcommand{\P}{\mathbb{P}}
\renewcommand{\E}{\mathbb{E}}

\def\R{{\mathbb R}}

\def\S{{\bf S}}

\def\x{{\bf x}}

\def\0{{\bf 0}}
\def\1{{\bf 1}}

\def\NM{{\mathcal N}}

\def\EB{{\mathbb E}}
\def\RB{{\mathbb R}}
\def\PB{{\mathbb P}}

\def\argmin{\mathop{\rm argmin}}

\def\var{\mathsf{var}}

\def\tr{\mathrm{tr}}

\def\ve{\varepsilon}

\renewcommand{\S}{\mathcal{S}}
\renewcommand{\L}{\mathcal{L}}

\newcommand{\opt}{\textup{opt}}
\newcommand{\ttop}{^{\top}}
\newcommand{\tsum}{\textstyle\sum}
\newcommand{\ts}{\textstyle}

\newcommand*{\qed}{\hfill\ensuremath{\square}}%

\begin{document}
\title{Error Estimation for Randomized Least-Squares Algorithms via the Bootstrap\thanks{A 9-page short version has appeared in the International Conference on Machine Learning (ICML), 2018.}}

\author{\name  Miles E.~Lopes \\
        \addr melopes@ucdavis.edu \\
            Department of Statistics\\
            University of California at Davis \\
            Davis, CA 95616, USA \\
        \AND
        \name Shusen Wang  \\
        \addr wssatzju@gmail.com \\
            Department of Computer Science \\
            Stevens Institute of Technology \\
            Hoboken, NJ 07030, USA \\
        \AND
        \name Michael W.\ Mahoney  \\
            \addr mmahoney@stat.berkeley.edu \\
            International Computer Science Institute and Department of Statistics \\
            University of California at Berkeley \\
            Berkeley, CA 94720, USA 
        }
        

\maketitle

\begin{abstract}%
Over the course of the past decade, a variety of randomized algorithms have been proposed for computing approximate least-squares (LS) solutions in large-scale settings. A longstanding practical issue is that, for any given input, the user rarely knows the \emph{actual error} of an approximate solution (relative to the exact solution). Consequently, the user often appeals to worst-case error bounds that tend to offer only qualitative guidance.
As a more practical alternative, we propose a bootstrap method to compute  \emph{a posteriori error estimates} for randomized LS algorithms. These estimates permit the user to numerically assess the error of a given solution, and to predict how much work is needed to improve a ``preliminary'' solution. From a practical standpoint, the method also has considerable flexibility, insofar as it can be applied to several popular sketching algorithms, as well as a variety of error metrics. Moreover, the extra step of error estimation does not add much cost to an underlying sketching algorithm. Finally, we demonstrate the effectiveness of the method with both theoretical and empirical results.
\end{abstract}


\section{Introduction}

Randomized sketching algorithms have been intensively studied in recent years as a general approach to computing fast approximate solutions to large-scale least-squares (LS) problems~\citep{drineas2006sampling,rokhlin2008fast,avron2010blendenpik,drineas2011faster,mahoney2011ramdomized,drineas2012fast,clarkson2013low,woodruff2014sketching,ma2014statistical,meng2014,pilanci2015,pilanci2016iterative}. During this time, much progress has been made in analyzing the performance of these algorithms, and  existing theory provides a good qualitative description of  approximation error (relative to the exact solution) in terms of various problem parameters. However, in practice, the user rarely knows the \emph{actual error} of a randomized solution, or how much extra computation may be needed to achieve a desired level of accuracy.

A basic source of this problem is that it is difficult to translate theoretical error bounds into numerical error bounds that are tight enough to be quantitatively meaningful. 
For instance, theoretical bounds are often formulated to hold for the worst-case input among a large class of possible inputs.
Consequently, they are often pessimistic for ``generic'' problems, and they may not account for the structure that is unique to the input at hand.
Another practical issue is that these bounds typically involve constants that are either conservative, unspecified, or dependent on unknown parameters.

In contrast with worst-case error bounds, we are interested in ``a posteriori'' error estimates.
By this, we mean error bounds that can be estimated numerically in terms of the computed solution or other observable information.
Although methods for obtaining a posteriori error estimates are well-developed in some areas of computer science and applied mathematics, there has been very little development for randomized sketching algorithms (cf.~Section~\ref{sec:relatedwork}).
(For brevity, we will usually omit the qualifier `a posteriori' from now on when referring to error estimation.)

The main purpose of this paper is to show that it is possible to directly estimate the error of randomized LS solutions in a way that is both practical and theoretically justified.
Accordingly, we propose a flexible estimation method that can enhance existing sketching algorithms in a variety of ways. 
In particular, we will explain how error estimation can help the user to (1) select the ``sketch size'' parameter, (2) assess the convergence of iterative sketching algorithms, and (3) measure error in a wider range of metrics than can be handled by existing theory.

\subsection{Setup and Background}

Consider a large, overdetermined LS problem, involving a rank $d$ matrix $A\in\RB^{n\times d}$, and a vector $b\in\RB^{n}$, where $n \gg d$. These inputs are viewed as deterministic, and the exact solution is denoted
\begin{equation}\label{eqn:orig}
x_{\opt}:=\argmin_{x\in\RB^d}\|Ax-b\|_2.
\end{equation}

The large number of rows $n$ is often a major computational bottleneck, and sketching algorithms overcome this obstacle by effectively solving a smaller problem involving $m$ rows, where  $d\ll m\ll n$. In general, this reduction is carried out with a random sketching matrix $S\in\RB^{m\times n}$ that maps the full matrix $A$ into a smaller sketched matrix $\tilde{A} := SA$ of size $m\times d$. However, various sketching algorithms differ in the way that the matrix $S$ is generated, or the way that $\tilde A$ is used. Below, we quickly summarize three of the most well-known types of sketching algorithms for LS.

\textbf{Classic Sketch (CS).} \ \
For a given sketching matrix $S$, this type of algorithm produces a solution 
\begin{equation}\label{eqn:CS}
\tilde x:=\argmin_{x\in\RB^d}\ts \|S(Ax-b)\|_2,
\end{equation}
and chronologically, this was the first type of sketching algorithm for LS~\citep{drineas2006sampling}.

\textbf{Hessian Sketch (HS).}  \  \
The HS algorithm modifies the objective function in the problem~\eqref{eqn:orig} so that its Hessian is easier to compute~\citep{pilanci2016iterative,Becker:2017}, leading to a solution 
\begin{equation}\label{eqn:HS}
\breve x:=\underset{x\in\RB^d}{\argmin}\Big\{\ts\frac{1}{2}\|SAx\|_2^2-\langle  A\ttop b, \, x\rangle \Big\}.
\end{equation}
This algorithm is also called ``partial sketching''.

\textbf{Iterative Hessian Sketch (IHS).} \ \
One way to extend HS is to refine the solution iteratively. For a given iterate \smash{$\hat x_i\in\RB^d$,} the following update rule is used
$$\hat x_{i+1}:=\underset{x\in\RB^d}{\argmin}\Big\{\ts\frac{1}{2}\|S_{i+1} A(x-\hat x_{i})\|_2^2+\langle A\ttop (A\hat x_i-b)\, , \, x\rangle \Big\},$$
where $S_{i+1}\in\RB^{m\times n}$ is a random sketching matrix that is generated independently of $S_1,\dots,S_i$, as proposed in~\cite{pilanci2016iterative}. 
If we let $t\geq 1$ denote the total number of IHS iterations, then we will generally write $\hat x_t$ to refer to the final output of IHS.

\textbf{Remark.} 
If the initial point for IHS is chosen as $\hat x_0=0$, then the first iterate $\hat x_1$ is equivalent to the HS solution $\breve x$ in equation~\eqref{eqn:HS}.
Consequently, HS may be viewed as a special case of IHS, and so we will restrict our discussion to CS and IHS for simplicity.

With regard to the choice of the sketching matrix, many options have been considered in the literature, and we refer to the surveys~\cite{mahoney2011ramdomized} and~\cite{woodruff2014sketching}. Typically, the matrix $S$ is generated so that the relation $\EB[S\ttop S]=I_n$ holds, and that the rows of $S$ are i.i.d.~random vectors (or nearly i.i.d.). Conceptually, our proposed method only relies on these basic properties of $S$, and in practice, it can be implemented with any sketching matrix.

To briefly review the computational benefits of sketching algorithms, first recall that the cost of solving the full least-squares problem~\eqref{eqn:orig} by standard methods is $\mathcal{O}(nd^2)$~\citep{Golub}. 
On the other hand, if the cost of computing the matrix product $SA$ is denoted $C_{\text{sketch}}$, and if a standard method is used to solve the sketched problem~\eqref{eqn:CS}, then the total cost of CS is $\mathcal{O}(md^2+C_{\text{sketch}})$.  
Similarly, the total cost of IHS with $t$ iterations is $\mathcal{O}(t(md^2+C_{\text{sketch}}))$.
Regarding the sketching cost $C_{\text{sketch}}$, it depends substantially on the choice of $S$, but there are many types that improve upon the naive $\mathcal{O}(mnd)$ cost of unstructured matrix multiplication. For instance, if $S$ is chosen to be a Sub-sampled Randomized Hadamard Transform (SRHT), then $C_{\text{sketch}}=\mathcal{O}(nd\log(m))$~\citep{ailon2006approximate,sarlos2006improved,ailon2009fast}. Based on these considerations, sketching algorithms can be more efficient than traditional LS algorithms when $md^2+nd\log(m)\ll nd^2$.

\subsection{Problem Formulation} 

For any problem instance, we will estimate the errors of the random vectors $\tilde x$ and $\hat x_t$  in terms of high-probability bounds. Specifically, 
if we let $\|\cdot\|_{\circ}$ denote \emph{any} norm on $\RB^d$,
and let $\alpha\in(0,1)$ be fixed, then our goal is to construct numerical estimates $\tilde\ve(\alpha)$ and $\hat \ve_t(\alpha)$, such that the bounds 
\begin{align}
\|\tilde x - x_{\opt}\|_{\circ} & \leq \tilde \ve(\alpha)\label{eqn:csbound}\\[0.3cm]
\|\hat x_t - x_{\opt}\|_{\circ}& \leq \hat \ve_t(\alpha) \label{eqn:ihsbound}
\end{align}
each hold with probability at least $1-\alpha$. (This probability will account for the randomness in both the sketching algorithm, and the bootstrap sampling described below.)
Also,
the algorithm for computing $\tilde \ve(\alpha)$ or $\hat\ve_t(\alpha)$ should be efficient enough so that the total cost of computing $(\tilde x,\tilde \ve(\alpha))$ or $(\hat x_t,\hat \ve_t(\alpha))$ is still much less than the cost of computing the exact solution $x_{\opt}$ --- otherwise, the extra step of error estimation would defeat the purpose of sketching. (This cost will be addressed in Section~\ref{sec:computation}.)
Since $x_{\opt}$ is unknown to the user, it might seem surprising that it is possible to construct error estimates that satisfy the conditions above,
and indeed, the limited knowledge of $x_{\opt}$ is the main source of difficulty.

\subsection{Main Contributions}
\label{sec:contrib}
At a high level, a distinguishing feature of our approach is that it applies inferential ideas from statistics in order to enhance large-scale computations. To be more specific, the novelty of this approach is that it differs from the traditional framework of using bootstrap methods to quantify uncertainty arising from data~\citep{Davison}. 
Instead, we are using these methods to quantify uncertainty in the outputs of randomized algorithms --- and there do not seem to be many works that have looked at the bootstrap from this angle.
From a more theoretical standpoint, another main contribution is that we offer the first guarantees for a posteriori error estimation involving the CS and IHS algorithms. (As a clarification, it should be noted that these results appeared in a conference version of the current work~\citep{LopesICML:2018}, but the proofs here have not previously been published.)

Looking beyond the present setting, there may be further opportunities for using bootstrap methods to estimate the errors of other randomized algorithms. In concurrent work, we have taken this approach in the distinct settings of randomized matrix multiplication, and randomized ensemble classifiers~\citep{LWM2017,L2018}.

\subsection{Related work}
\label{sec:relatedwork}

The general problem of error estimation for approximation algorithms has been considered in a wide range of situations, and we refer to the following works for surveys and examples: \cite{Pang:1987,Verfurth,Jiranek,Ainsworth,Colombo}. In the context of sketching algorithms, there is only a handful of papers that address error estimation, and these are geared toward low-rank approximation~\citep{Liberty:2007,Woolfe:2008,Halko:2011}, or matrix multiplication~\citep{sarlos2006improved,LWM2017}. 
In addition to the works just mentioned,
the recent preprint~\cite{Ahfock} explores statistical properties of the CS and HS algorithms, and it develops analytical formulas for describing how $\tilde x$ and $\breve x$ fluctuate around $x_{\opt}$.
Although these formulas offer insight into error estimation, their application is limited by the fact that they involve unknown parameters. Also, the approach in~\citep{Ahfock} does not address IHS. Lastly, error estimation for LS approximations can be studied from a Bayesian perspective, and this has been pursued in the paper~\citep{Bartels2016}, but with a focus on algorithms that differ  from the ones studied here.

\textbf{Notation.} 
The following notation is needed for our proposed algorithms. Let $\tilde b:=Sb\in\RB^m$ denote the sketched version of $b$. If $\textbf{i}=(i_1,\dots,i_m)$ is a vector containing $m$ numbers from $\{1,\dots,m\}$, then $\tilde A(\textbf{i},:)$ refers to the $m\times d$ matrix whose $j$th row is equal to the $i_j$th row of $\tilde A$. Similarly, the $j$th component of the vector $\tilde b(\textbf{i})$ is the $i_j$th component of $\tilde b$. Next, for any fixed $\alpha\in (0,1)$, and any finite set of real numbers $C=\{c_1,\dots,c_k\}$, the expression $\text{quantile}(c_1,\dots,c_k;1-\alpha)$ is defined as the smallest element $c_{i_0}\in C$ for which the sum $\frac{1}{k} \sum_{i=1}^k 1\{c_i\leq c_{i_0}\}$ is at least $1-\alpha$. Lastly, the distribution of a random variable $U$ is denoted $\mathcal{L}(U)$, and the conditional distribution of $U$ given a random variable $V$ is denoted $\mathcal{L}(U|V)$.

\section{Method}
The proposed bootstrap method is outlined in Sections~\ref{sec:bootCS} and~\ref{sec:bootIHS} for the cases of CS and IHS respectively. The formal analysis can be found in the proof of Theorem~\ref{thm:main} in the appendices. Later on, in Section~\ref{sec:computation}, we discuss computational cost and speedups.

\subsection{Error Estimation for CS}\label{sec:bootCS}

The main challenge we face is that the distribution of the random variable $\|\tilde x -x_{\opt}\|_{\circ}$ is unknown. If we had access to this distribution, we could find the tightest possible upper bound on $\|\tilde x -x_{\opt}\|_{\circ}$ that holds with probability at least $1-\alpha$. (This bound is commonly referred to as the $(1-\alpha)$-quantile of the random variable $\|\tilde x-x_{\opt}\|_{\circ}$.)

From an intuitive standpoint, the idea of the proposed bootstrap method is to artificially generate many samples of a random vector, say $\tilde x^*$, whose fluctuations around $\tilde x$ are statistically similar to the fluctuations of $\tilde x$ around $x_{\opt}$. 
In turn, we can use the empirical $(1-\alpha)$-quantile of the values $\|\tilde x^*-\tilde x\|_{\circ}$ to obtain the desired estimate $\tilde \ve(\alpha)$ in~\eqref{eqn:csbound}.

\textbf{Remark.} 
As a technical clarification, it is important to note that our method relies only on a \emph{single run} of CS, involving just one sketching matrix $S$.
Consequently, 
the bootstrapped vectors $\tilde x^*$ will be generated conditionally on the given $S$. In this way, the bootstrap  aims to generate random vectors $\tilde x^*$, such that for a given draw of $S$, the conditional distribution $\mathcal{L}(\tilde x^*-\tilde x\,|\,S)$ is approximately equal to the unknown distribution \smash{$\mathcal{L}(\tilde x-x_{\opt})$.}

\begin{mdframed}
	\begin{alg} \label{alg:cs}
		{\bf (Error estimate for CS).}\\
		{\bf Input:} A positive integer $B$, and the sketches $\tilde A$, $\tilde b$, and $\tilde x$.\\[0.2cm]
		{\bf For } $l=1,\dots,B$\; {\bf do} 
		\begin{enumerate}
			\item Draw a vector $\textbf{i}:=(i_1,\dots,i_m)$ by sampling $m$ numbers with replacement from $\{1,\dots,m\}$.
			\item Form the matrix $\tilde A^*:=\tilde A(\textbf{i},:)$, and vector $\tilde b^*:=\tilde b(\textbf{i})$. 
			\item Compute the following vector and scalar,
			\begin{equation}\label{eqn:tildexalg}
			\tilde  x^* := \argmin_{x\in\RB^d}\|\tilde A^*x-\tilde b^*\|_2 \text{ \ \  \ \ \ and \ \ \ \ \  } \ve_l^*:=\|\tilde x^*-\tilde x\|_{\circ}.
			\end{equation}
		\end{enumerate}
		{\bf Return:} 
		$\tilde \ve(\alpha):=\text{quantile}(\ve_1^*,\dots,\ve_B^*; 1-\alpha)$.
	\end{alg}
\end{mdframed}

\textbf{Heuristic interpretation of Algorithm~\ref{alg:cs}.} To explain why the bootstrap works,  let $\mathbb{S}_A$ denote the set of positive semidefinite matrices  \smash{$M\in\RB^{n\times n}$} such that $A\ttop MA$ is invertible, and define the map \smash{$\psi:\mathbb{S}_A\to\RB^d$} according to
\begin{equation}
\psi(M)=(A\ttop M A)^{-1}A\ttop M b.
\end{equation}
This map leads to the relation\footnote{For standard types of sketching matrices,  the event \smash{$S\ttop S \in\mathbb{S}_A$} occurs with high probability when $A\ttop A$ is invertible and $m$ is sufficiently larger than $d$ (and similarly for $S^{*\top}S^*$).
}\label{Sfootnote}
\begin{equation}\label{eqn:relation}
\tilde x -x_{\opt} = \psi(S\ttop S)-\psi(I_n),
\end{equation}
where $I_n$ denotes the $n\times n $ identity matrix. By analogy, if we let $S^*\in\RB^{m\times n}$ denote a matrix obtained by sampling $m$ rows from $S$ with replacement, then $\tilde x^*$ can be written as
\begin{equation}\label{eqn:tildexstar}
\tilde x^*=\argmin_{x\in\RB^d}  \|S^*(Ax-b)\|_2,
\end{equation}
and the definition of $\psi$ gives 
\begin{equation}\label{eqn:bootrelation}
\tilde x^*-\tilde x \  = \ \psi(S^{*\top} S^*) - \psi(S\ttop S).
\end{equation}

Using the corresponding relations~\eqref{eqn:relation} and~\eqref{eqn:bootrelation}, it becomes easier to explain why the distributions $\mathcal{L}(\tilde x-x_{\opt})$ and \smash{$\mathcal{L}(\tilde x^*-\tilde x|S)$} should be nearly equal.

To proceed, if we let $s_1,\dots,s_m\in\RB^n$ denote the rows of $\sqrt{m}S$, it is helpful to note the basic algebraic fact
\begin{equation}
S\ttop S-I_n= \ts\frac 1m \tsum_{i=1}^m (s_is_i\ttop -I_n).
\end{equation}
Given that sketching matrices are commonly constructed so that $s_1,\dots,s_m$ are i.i.d.~(or nearly i.i.d.) with $\EB[s_1s_1\ttop]=I_n$, the matrix $S\ttop S$ becomes an increasingly good approximation to $I_n$ as $m$ becomes large. Hence, it is natural to consider a first-order expansion of the right side of~\eqref{eqn:relation}, 
\begin{equation}\label{eqn:firstorder}
\tilde x -x_{\opt} \ \approx \  \psi_{I_n}'(S\ttop S-I_n),
\end{equation}
where $\psi'_{I_n}$ is the differential of the map $\psi$ at $I_n$. 
Likewise, if we define a set of vectors $v_1,\dots,v_m\in\RB^d$ as \smash{$v_i:=\psi_{I_n}'(s_is_i\ttop-I_n)$}, then the linearity of $\psi_{I_n}'$ gives
\begin{equation}\label{eqn:sampleav}
\tilde x -x_{\opt}\ \approx \ \ts\frac 1m \tsum_{i=1}^m v_i,
\end{equation}
and furthermore, the vectors $v_1,\dots,v_m$ are i.i.d.~whenever the vectors $s_1,\dots,s_m$ are.
Consequently, as the sketch size $m$ becomes large, the central limit theorem suggests that the difference $\sqrt m(\tilde x-x_{\opt})$ should be approximately Gaussian, 
\begin{equation}\label{eqn:cltmain}
\mathcal{L}\big(\sqrt m(\tilde x-x_{\opt})\big) \  \approx \ \mathcal{N}\big(0,\Sigma\big),
\end{equation}
where we put $\Sigma:=\EB[v_1v_1\ttop]$.

To make the connection with $\tilde x^*-\tilde x$, each of the preceding steps can be carried out in a corresponding manner. Specifically, if the differential of $\psi$ at $S\ttop S$ is sufficiently close to the differential at $I_n$, then an expansion of equation~\eqref{eqn:bootrelation} leads to the bootstrap analogue of~\eqref{eqn:sampleav},
\begin{equation}
\tilde x^*-\tilde x \ \approx \  \ts\frac 1m \tsum_{i=1}^m v_i^*,
\end{equation}
where $v_i^*:=\psi_{I_n}'(s_i^*s_i^{*\top}-S\ttop S)$, and the vector $s_i^{*}$ is the $i$th row of $\sqrt m S^*$. Since the row vectors $s_1^{*},\dots,s_m^{*}$ are obtained by sampling with replacement from $\sqrt m S$, it follows that the vectors $v_1^*,\dots,v_m^*$ are conditionally i.i.d.~given $S$, and also, $\EB[v_i^*|S]=0$. Therefore, if we condition on $S$, the central limit theorem suggests that as $m$ becomes large
\begin{equation}\label{eqn:bootcltmain}
\mathcal{L}\big(\sqrt m (\tilde x^*-\tilde x)\,|\,S\big) \  \approx \ \mathcal{N}\big(0,\hat\Sigma\big),
\end{equation}
where the conditional covariance matrix is denoted by $\hat\Sigma :=\EB\big[v_i^*v_i^{*\top}\big|S\big]$. Comparing the Gaussian approximations~\eqref{eqn:cltmain} and~\eqref{eqn:bootcltmain}, this heuristic argument indicates that the distributions $\mathcal{L}(\tilde x-x_{\opt})$ and $\mathcal{L}(\tilde x^*-\tilde x|S)$ should be close as long as $\hat \Sigma$ is close to  $\Sigma$, and when $m$ is large, this is enforced by the law of large numbers.

\subsection{Error Estimation for IHS}\label{sec:bootIHS}

At first sight, it might seem that applying the bootstrap to IHS would be substantially different than in the case of CS --- given that IHS is an iterative algorithm, whereas CS is a ``one-shot'' algorithm. However,  the bootstrap only needs to be modified slightly.
Furthermore, the bootstrap relies on just the final two iterations of \emph{a single run} of IHS.

To fix some notation, recall that $t$ denotes the total number of IHS iterations, and let $S_{t}\in\RB^{m\times n}$ denote the sketching matrix used in the last iteration. Also  define the matrix \smash{$\tilde A_{t}:=S_{t}A$,} and the gradient vector \smash{$g_{t-1}:=A\ttop(A\hat x_{t-1}-b)$} that is computed during the second-to-last iteration of IHS. Lastly, we note that the user can select any initial point $\hat x_0$ for IHS, and this choice does not restrict our method.

\begin{mdframed}
	\begin{alg} \label{alg:ihs}
		{\bf (Error estimate for IHS).}\\
		{\bf Input:} A positive integer $B$, the sketch $\tilde A_{t}$, the gradient $g_{t-1}$, the second-to-last iterate $\hat x_{t-1}$, and the last iterate $\hat x_t$. \\[0.2cm]
		{\bf For } $l=1,\dots,B$\; {\bf do} 
		\begin{enumerate}
			\item Draw a vector $\textbf{i}:=(i_1,\dots,i_m)$ by sampling $m$ numbers with replacement from $\{1,\dots,m\}$.
			\item Form the matrix $\tilde A_{t}^*:=\tilde A_{t}(\textbf{i},:)$.
			\item Compute the following vector and scalar,
			\begin{small}
			\begin{equation}\label{eqn:hatxalg}
			\hat x_t^* := \underset{x\in\RB^d}{\argmin} \Big\{\ts\frac{1}{2}\|\tilde A_{t}^*(x-\hat x_{t-1})\|_2^2+\big\langle g_{t-1}\, , \, x\big\rangle\Big\} \text{ \ \  \ \ \ and \ \ \ \ \  } \ve_{t,l}^*:=\|\hat x_t^*-\hat x_t\|_{\circ}.
			\end{equation}
		\end{small}%
		\end{enumerate}
		{\bf Return:} 
		$\hat\ve_t(\alpha):=\text{quantile}(\ve_{t,1}^*,\dots,\ve_{t,B}^*; 1-\alpha)$.
	\end{alg}
\end{mdframed}

\textbf{Remark.}
The ideas underlying the IHS version of the bootstrap are broadly similar to those discussed for the CS version. However, the details of this argument are much more involved than in the CS case, owing to the iterative nature of IHS. A formal analysis may be found in the proof of Theorem~\ref{thm:main} in the appendices.

\subsection{Computational Cost and Speedups}\label{sec:computation}

Of course, the quality control that is provided by error estimation does not come for free. Nevertheless, there are several special properties of Algorithms \ref{alg:cs} and \ref{alg:ihs} that keep their computational cost in check, and in particular, the cost of computing $(\tilde x, \tilde \ve(\alpha))$ or $(\hat x_t, \hat\ve_t(\alpha))$ is much less than the cost of solving the full LS problem~\eqref{eqn:orig}. These properties are summarized below.
\begin{enumerate}
	\item \textbf{Cost of error estimation is independent of $n$.}\\
	The inputs to Algorithms \ref{alg:cs} and \ref{alg:ihs} consist of pre-computed matrices of size $m\times d$, or pre-computed vectors of dimension $d$. Consequently, both algorithms are highly scalable in the sense that their costs do not depend on the large dimension $n$. As a point of comparison, it should be noted that sketching algorithms for LS generally have costs that scale linearly with $n$.
	\item \textbf{Implementation is embarrassingly parallel.}\\
	Due to the fact that each bootstrap sample is computed independently of the others, the for-loops in Algorithms \ref{alg:cs} and \ref{alg:ihs} can be easily distributed. \emph{ Furthermore, it turns out that in practice, as few as $B=20$ bootstrap samples are often sufficient to obtain good error estimates, as illustrated in Section~\ref{sec:experiment}}. Consequently, even if the error estimation is done on a single workstation, it is realistic to suppose that the user has access to $N$ processors such that the number of bootstrap samples per processor is $B/N=\mathcal{O}(1)$. If this is the case, and if $\|\cdot\|_{\circ}$ is any $\ell_p$ norm on $\mathbb{R}^d$, then it follows that the per-processor cost of both algorithms is only $\mathcal{O}(md^2)$. Lastly, the communication costs in this situation are also very modest. In fact, it is only necessary to send a single $m\times d$ matrix, and at most three $d$-vectors to each processor. In turn, when the results are aggregated, only $B$ scalars are sent back to the central processor.
	\item \textbf{{Bootstrap computations have free warm starts.}}\\
	The bootstrap samples $\tilde x^*$ and $\hat x_t^*$ can be viewed as perturbations of the actual sketched solutions $\tilde x$ and $\hat x_t$. This is because the associated optimization problems only differ with respect to resampled versions of $\tilde A$ and $\tilde b$. Therefore, if a sub-routine for computing  $\tilde x^*$ or $\hat x_t^*$ relies on an initial point, then $\tilde x$ or $\hat x_t$ can be used as warm starts at no additional cost. By contrast, note that warm starts are not necessarily available when $\tilde x$ or $\hat x_t$ are first computed. \emph{In this way, the computation of the bootstrap samples is easier than a naive repetition of the underlying sketching algorithm}.
	\item \textbf{Error estimates can be extrapolated.}\\
	The basic idea of extrapolation is to estimate the error of a ``rough'' initial sketched solution, say $\tilde{x}_{\text{init}}$ or $\hat{x}_{\text{init}}$, and then predict how much additional computation should be done to obtain a better solution $\tilde x$ or $\hat x_t$. There are two main benefits of doing this. First, the computation is \emph{adaptive}, in the sense that ``just enough'' work is done to achieve the desired degree of error. Secondly, when error estimation is based on the rough initial solutions $\tilde x_{\text{init}}$ or $\hat x_{\text{init}}$, the bootstrap computations are substantially faster, because $\tilde x_{\text{init}}$ and $\hat x_{\text{init}}$ are constructed from very small sketching matrices. There are also two ways that extrapolation can be done --- either with respect to the sketch size $m$, or the number of iterations $t$, and these techniques are outlined in the following paragraphs.
\end{enumerate}

\subsection{Extrapolating with respect to $m$ for CS}
The reasoning given in Section~\ref{sec:bootCS} based on the central limit theorem indicates that the standard deviation of \smash{$\|\tilde x  - x_{\opt}\|_{\circ}$} scales like $1/\sqrt{m}$ as a function of $m$. Therefore, if a rough initial solution $\tilde x_{\text{init}}$ is computed with a small sketch size $m_0$ satisfying $d<m_0<m$, then the fluctuations of \smash{$\|\tilde x_{\text{init}}  - x_{\opt}\|_{\circ}$} should be larger than those of $\|\tilde x-x_{\opt}\|_{\circ}$ by a factor of $\sqrt{m/m_0}$.  
This simple scaling relationship is useful to consider, because if we let $\tilde \ve_{\text{init}}(\alpha)$ denote the error estimate obtained by applying Algorithm \ref{alg:cs} to $\tilde x_{\text{init}}$, then  it is natural to expect that the re-scaled quantity 
\begin{equation}\label{eqn:basicextrap}
\tilde \ve_{\text{extrap},m}(\alpha)
\; := \; \sqrt{\ts\frac{m_0}{m}} \: \tilde\ve_{\text{init}}(\alpha)
\end{equation}  
should be approximately equal to the ordinary estimate $\tilde \ve(\alpha)$ for $\tilde x$. The advantage of $\tilde \ve_{\text{extrap},m}(\alpha)$ is that it is cheaper to compute, since the bootstrapping can be done with a \smash{$m_0\times d$} matrix, rather than an $m\times d$ matrix. Furthermore, once $\tilde \ve_{\text{init}}(\alpha)$ has been computed, the user can instantly obtain $\tilde{\ve}_{\text{extrap},m}(\alpha)$ as a function of $m$ for all $m>m_0$, using the scaling rule~\eqref{eqn:basicextrap}. In turn, this allows the user to ``look ahead'' and see how large $m$ should be chosen to achieve a desired level of accuracy. Simulations demonstrating the effectiveness of this technique are given in Section~\ref{sec:experiment}.

\subsection{Extrapolating with respect to $t$ for IHS}

The IHS algorithm is known to enjoy linear convergence in the $\ell_2$-norm under certain conditions~\citep{pilanci2016iterative}. This means that the $i$th iterate $\hat x_i$  satisfies the following bound with high probability
\begin{equation}\label{eqn:ihsextrap}
\|\hat x_i-x_{\opt}\|_{2} \leq  c\,\eta^i,
\end{equation}
where $c>0$ and $\eta\in (0,1)$ are unknown parameters that do not depend on $i$.

The simple form of this bound lends itself to extrapolation. Namely, if estimates $\hat c$ and $\hat \eta$ can be obtained after the first 2 iterations of IHS,
then the user can construct the extrapolated error estimate
\begin{equation}\label{eqn:ihsextrapsecond}
\hat \ve_{\text{extrap},i}(\alpha):= \hat c \,\hat\eta^i,
\end{equation}
which predicts how the error will decrease at all subsequent iterations $i\geq 3$.
As a result, the user can adaptively determine how many extra iterations (if any) are needed for a specified error tolerance.
Furthermore, with the help of Algorithm \ref{alg:ihs}, it is straightforward to estimate $c$ and $\eta$. Indeed, from looking at the condition~\eqref{eqn:ihsextrap}, we desire estimates $\hat c$ and $\hat \eta$ that solve the two equations
\begin{equation}
\hat c\,\hat \eta = \hat \ve_1(\alpha) \ \ \ \text{ and }  \ \ \ \ \hat c\,\hat \eta^{2}=\hat\ve_{2}(\alpha),
\end{equation}
and direct inspection shows that the choices
$\hat \eta:=\ts\frac{\hat \ve_{2}(\alpha)}{\hat\ve_1(\alpha)}$
and $\hat c :=\ts\frac{\hat\ve_1(\alpha)}{\hat\eta}$
serve this purpose. In Section~\ref{sec:experiment}, our experiments show that this simple extrapolation procedure works remarkably well.

\section{Main Result}\label{sec:main}
In this section, we show that the estimates $\tilde \ve(\alpha)$ and $\hat \ve_t(\alpha)$ are consistent --- in the sense that they satisfy the desired conditions~\eqref{eqn:csbound} and~\eqref{eqn:ihsbound} as the problem size becomes large. The setup and assumptions for our main result are given below.

\textbf{Asymptotics.} 
We consider an asymptotic framework involving a sequence of LS problems indexed by $n$. This means that we allow each of the objects $A=A(n), S=S(n),$ and $b=b(n)$ to implicitly depend on $n$. Likewise, the solutions $\tilde x=\tilde x(n)$ and $\hat x_t=\hat x_t(n)$ implicitly depend on $n$.

Since sketching algorithms are most commonly used when $d\ll n$, our results will treat $d$ as fixed while $n\to\infty$. Also, the sketch size $m$ is often selected as a large multiple of $d$, and so we treat $m=m(n)$ as diverging simultaneously with $n$. However, we make no restriction on the size of the ratio $m/n$, which may tend to 0 at any rate. In the same way, the number of bootstrap samples $B=B(n)$ is assumed to diverge with $n$, and the ratio $B/n$ may tend to 0 at any rate. With regard to the number of iterations $t$, its dependence on $n$ is completely unrestricted, and $t=t(n)$ is allowed to remain fixed or diverge with $n$. (The fixed case with $t=1$ is of interest since it describes the HS algorithm.) Apart from these scaling conditions, we use the following two assumptions on $A$ and $b$, as well as the sketching matrices.

\textbf{Assumption 1.} \emph{ The matrix \smash{$H_n:=\frac 1n A\ttop A$} is positive definite for each $n$, and there is a positive definite matrix \smash{$H_{\infty}\in\RB^{d\times d}$} such that \smash{$\sqrt m (H_n-H_{\infty})\to 0$} as \smash{$n\to\infty$}. Also, if $\texttt{g}_n:=\frac 1n A\ttop b$, then there is a vector $\texttt{g}_{\infty}\in\RB^d$ such that $\sqrt m(\texttt{g}_n-\texttt{g}_{\infty})\to 0$. Lastly, let $a_1,\dots,a_n$ denote the rows of $A$, and let $e_1,\dots,e_n$ denote the standard basis vectors in $\RB^n$. Then, for any fixed matrix $C\in\RB^{d\times d}$ and fixed vector $c\in\RB^d$, the sum $\ts\frac{1}{n^2}\sum_{j=1}^n \big(a_j\ttop Ca_j+ e_j\ttop b c\ttop a_j\big)^2$ converges to a limit (possibly zero) as $n\to\infty$.}

In essence, this assumption ensures that the sequence of LS problems is ``asymptotically stable'', in the sense that the optimal solution $x_{\opt}$ does not change erratically from $n$ to $n+1$.

\textbf{Assumption 2.} \emph{  In the case of CS, the rows of $S$ are generated as i.i.d.~vectors, where the $i$th row is of the form $\frac{1}{\sqrt m} (s_{i,1},\dots,s_{i,n})$, and the random variables  $s_{i,1},\dots,s_{i,n}$ are i.i.d.~with mean 0, variance 1, $\EB[s_{1,1}^4]>1$, and $\EB[s_{1,1}^8]<\infty$.
	In addition, the distribution of $s_{1,1}$ remains fixed with respect to $n$, and in the case of IHS, the matrices $S_1,\dots, S_t$ are i.i.d.~copies of $S$.}

\textbf{Remarks.} \ To clarify the interpretation of our main result, it is important to emphasize that $A$ and $b$ are viewed as deterministic, and the probability statements arise only from the randomness in the sketching algorithm, and the randomness in the bootstrap sampling. From an operational standpoint, the result says that as the problem size becomes large $(n\to\infty)$, the outputs $\tilde \ve(\alpha)$ and $\hat\ve_t(\alpha)$ of our method will bound the errors $\|\tilde x-x_{\opt}\|_{\circ}$ and $\|\hat x_t-x_{\opt}\|_{\circ}$ with a probability that is effectively $1-\alpha$ or larger.
\begin{theorem} \label{thm:main}
	Let $\|\cdot\|_{\circ}$ be any norm on $\RB^d$, and suppose that Assumptions 1 and 2 hold. Also, for any number $\alpha\in (0,1)$ chosen by the user, let $\tilde \ve(\alpha)$ and  $\hat \ve_t(\alpha)$ be the outputs of Algorithms \ref{alg:cs} and \ref{alg:ihs} respectively. Then, there is a sequence of numbers $\delta_n>0$ satisfying $\delta_n\to 0$ as $n\to\infty$, such that the following inequalities hold for all $n$,
	\begin{equation}\label{eqn:cs_thm}
	\ts
	\PB\Big(\|\tilde x-x_{\opt}\|_{\circ} \leq \tilde \ve(\alpha)\Big) \geq  1-\alpha-\delta_n,
	\end{equation}
	~\\[-0.5cm]
	and\\[-0.5cm]
	\begin{equation}\label{eqn:ihs_thm}
	\ts
	\PB\Big(\|\hat x_t-x_{\opt}\|_{\circ} \leq \hat\ve_t(\alpha)\Big) \geq 1-\alpha-\delta_n.
	\end{equation}
\end{theorem}

\textbf{Remarks.} Although this result can be stated in a concise form, the proof is actually quite involved. Perhaps the most significant technical obstacle is the sequential nature of the IHS algorithm. To handle the dependence of $\hat x_t$ on the previous iterates, it is natural to analyze $\hat x_t$ conditionally on them. However, because the set of previous iterates can grow with $n$, it seems necessary to establish distributional limits that hold ``uniformly'' over those iterates --- and this need for uniformity creates difficulties when applying standard arguments.

More generally, to place this result within the context of the sketching literature, it is worth noting that guarantees for sketching algorithms typically show that a sketched solution is close to an exact solution with high probability (up to multiplicative constants). 
By contrast, Theorem~\ref{thm:main} is more fine-grained, since it is concerned with \emph{distributional approximation}, in terms of specific quantiles of the random variables \smash{$\|\tilde x-x_{\opt}\|_{\circ}$} or $\|\hat x_t-x_{\opt}\|_{\circ}$. 
In particular, the lower bounds are asymptotically equal to $1-\alpha$ and \emph{do not involve any multiplicative constants}. 
Lastly, it should also be noticed that the norm $\|\cdot\|_{\circ}$ is arbitrary, whereas other analyses of sketching algorithms are often restricted to particular norms.

\section{Experiments}\label{sec:experiment}
\def\std{\mathsf{std}}

In this section, we present experimental results in the contexts of CS and IHS. At a high level, there are two main takeaways: 
(1) The extrapolation rules accurately predict how estimation error depends on $m$ or $t$, and this is shown in a range of conditions.
(2) In all of the experiments, the algorithms are implemented with only $B=20$ bootstrap samples. The fact that favorable results can be obtained with so few samples underscores the point that the method incurs only modest cost in exchange for an accuracy guarantee.

\subsection{Data examples}

Our numerical results are based on four linear regression datasets; two natural, and two synthetic. The natural datasets
`YearPredictionMSD', $n=463,\!715$, $d=90$, abbrev.~MSD), and `cpusmall' ($n=8,192$, $d=12$, abbrev.~CPU) 
are available at the LIBSVM repository~\citep{libsvm}. 

The synthetic datasets are both of size $(n=50,\!000, d=100)$, but they differ with respect to the condition number of $A\ttop A$.  The condition numbers in the  `Ill-conditioned' and  `Well-conditioned' cases are respectively $10^{12}$ and $10^2$. 
To generate the matrix $A \in \RB^{n\times d}$,
we selected the factors of its s.v.d.~ $A = U \diag (\sigma ) V^T$ as follows:
\begin{itemize}
	\item 
	The rows of a matrix $X \in \RB^{n\times d}$ were sampled i.i.d.~from a multivariate $t$-distribution, $ t_{2}( \mu,C)$, with $2$ degrees of freedom, mean $\mu=0$, and covariance  $c_{ij}=2\times 0.5^{|i-j|}$ where $C=[c_{ij}]$.
	Let $U\in\R^{d\times d} $ be the orthogonal basis of $X$ (i.e.~the $Q$ factor in a QR decomposition of $X$). In this way, $A$ has high row coherence, which can create difficulties for sampling-based sketching matrices.
	\item
	We generated $V \in \R^{d\times d}$ as the orthogonal basis of a $d\times d$ standard  Gaussian matrix.
	\item
	We generated the vector $\sigma \in \RB^d$ in two ways.
	First, we let the entries of $c \in \R^{d}$ be equally spaced between $0$ and $-6$, and then we put $\sigma_i = 10^{c_i}$ for all $i\in [d]$.
	In this way, $A\ttop A$ has a condition number of $10^{12}$.
	Second, we let the entries of $\sigma$ be equally spaced between $0.1$ and $1$.
	In this way, the condition number of $A\ttop A$ is only $100$.
\end{itemize}
Lastly, we defined the approximately sparse vector $x = [\1_{0.2d} , \e \cdot \1_{0.6d}, \1_{0.2d}]\in\R^d$ with $\e=0.1$, and $\1_k$ being the all-ones vector of dimension $k$. Then, by generating a vector $z\in\RB^n$ whose coordinates are i.i.d.~$\NM (0, \tau^2)$ with $\tau=.001$, we put $b = A x + z$.
In previous work, a number of other experiments in randomized matrix computations have been designed in this way~\citep{ma2014statistical,yang2016implementing}.

\subsection{Experiments for CS.} 

\begin{figure*}[h]
	\begin{center}
		\includegraphics[width=0.99\textwidth]{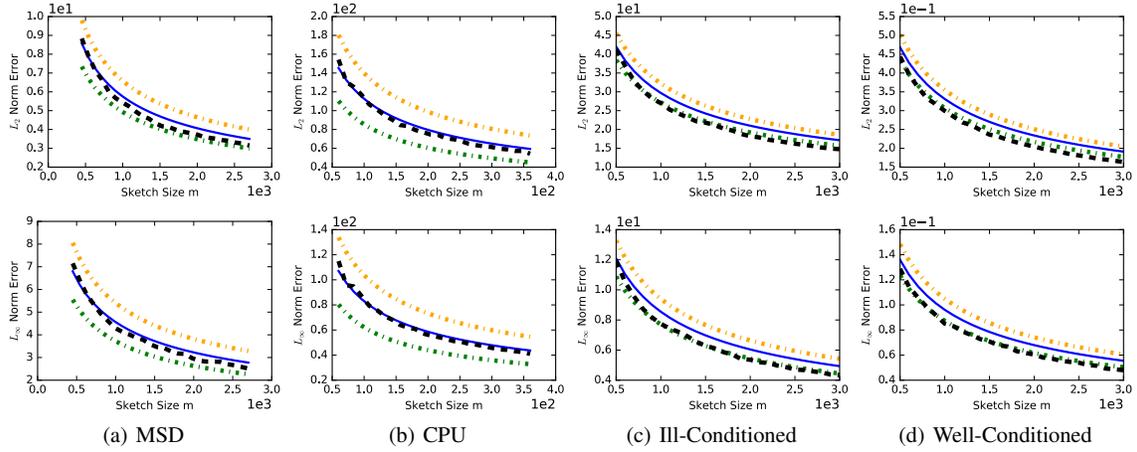}
	\end{center}
	\vspace{-4mm}
	\caption{\emph{Numerical results for CS with extrapolation.}
		The black dashed curve represents the ideal benchmark $\ve_{\text{CS},m}(.05)$ described in the text. The average extrapolated estimate is shown in blue, with the yellow and green curves being one standard deviation away.
		Note: The upper row shows results for $\ell_2$ error ($\|\cdot\|_{\circ}=\|\cdot\|_2$), and the lower row shows results for $\ell_{\infty}$ error ($\|\cdot\|_{\circ}=\|\cdot\|_{\infty}$).
	}
	\label{fig:cs}
	\vspace{-1mm}
\end{figure*}

For each value of $m$ in the grid $\{5d, \dots, 30d\}$, we  generated 1,000 independent SRHT sketching matrices \smash{$S\in\RB^{m\times n}$,} leading to 1,000 realizations of of $(\tilde A, \tilde b,\tilde x)$. 
Then,  we computed the .95 sample quantile among the 1,000 values of $\|\tilde x- x_{\opt}\|$ at each grid point. 
We denote this value as $\ve_{\text{CS},m}(.05)$, and we view it as an ideal benchmark that satisfies \smash{$\PB\big(\|\tilde x -x_{\opt}\|\leq \ve_{\text{CS},m}(.05)\big)\approx .95$} for each $m$. 
Also, the value $\ve_{\text{CS},m}(.05)$ is plotted as a function of $m$ with the dashed black line in Figure~\ref{fig:cs}. 
Next, using an initial sketch size of $m_0=5d$, we applied Algorithm \ref{alg:cs} to each of the 1,000 realizations of $\tilde A\in\RB^{m_0\times d}$ and $\tilde b\in\RB^{m_0}$ computed previously, leading to 1,000 realizations of the initial error estimate $\tilde \ve_{\text{init}}(.05)$. 
In turn, we applied the extrapolation rule~\eqref{eqn:basicextrap} to each realization of $\tilde \ve_{\text{init}}(.05)$, providing us with 1,000 extrapolated curves of $\tilde \ve_{\text{extrap},m}(.05)$ at all grid points $m\geq m_0$. The average of these curves is plotted in blue in Figure~\ref{fig:cs}, with the yellow and green curves being one standard deviation away.

\textbf{Comments on results for CS.} An important conclusion to draw from Figure~\ref{fig:cs} is that the extrapolated estimate $\tilde \ve_{\text{extrap},m}(.05)$ is a nearly unbiased estimate of $\tilde \ve_{\text{CS},m}(.05)$ at values of $m$ that are well beyond $m_0$.
This means that in addition to yielding accurate estimates, the extrapolation rule~\eqref{eqn:basicextrap} provides substantial computational savings --- because the bootstrap computations can be done at a value $m_0$ that is much smaller than the value $m$ ultimately selected for a higher quality $\tilde x$. Furthermore, these conclusions hold regardless of whether the error is measured with the $\ell_2$-norm $(\|\cdot\|_{\circ}=\|\cdot\|_2)$ or the $\ell_{\infty}$-norm ($\|\cdot\|_{\circ}=\|\cdot\|_{\infty}$), which correspond to the top and bottom rows of Figure~\ref{fig:cs}.

\begin{figure*}[h]
	\begin{center}
		\includegraphics[width=0.99\textwidth]{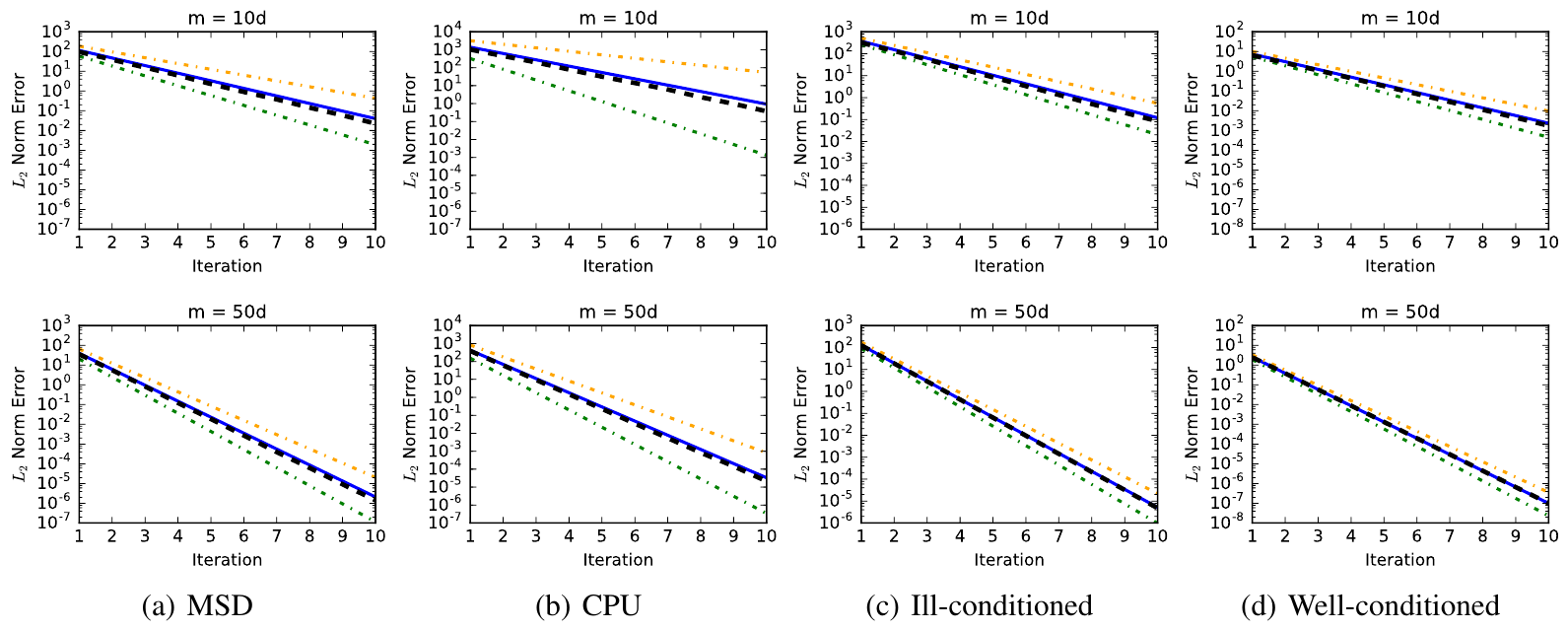}
	\end{center}
	\vspace{-4mm}
	\caption{\emph{Numerical results for IHS with extrapolation.}
		The black dashed curve represents the ideal benchmark $\ve_{\text{IHS},i}(.05)$ described in the text. The average extrapolated estimate is shown in blue, with the yellow and green curves being one standard deviation away. The upper row shows results for $m=10d$, and the lower row shows results for $m=50d$.
	}
	\label{fig:ihs}
	\vspace{-1mm}
\end{figure*}

\textbf{Experiments for IHS.} \ The experiments for IHS were organized similarly to the case of CS, except that the sketch size $m$ was fixed (at either $m=10d$, or $m=50d$), and results were considered as a function of the iteration number. 
To be specific, the IHS algorithm was run 1,000 times, with $t=10$ total iterations on each run, and with SRHT sketching matrices being used at each iteration. For a given run, the successive error values $\|\hat x_i-x_{\opt}\|_2$ at $i=1,\dots,10$, were recorded. 
At each $i$, we computed the .95 sample quantile among the 1,000 error values, which is denoted as $\ve_{\text{IHS},i}(.05)$, and is viewed as an ideal benchmark that satisfies $\PB\big(\|\hat x_i-x_{\opt}\|_2\leq  \ve_{\text{IHS},i}(.05)\big)\approx .95$. 
In the plots, the value $\ve_{\text{IHS},i}(.05)$ is plotted with the dashed black curve as a function of $i=1,\dots,10$. In addition, for each of the 1,000 runs, we applied Algorithm \ref{alg:ihs} at $i=1$ and $i=2$, producing 1,000 extrapolated values $\hat \ve_{\text{extrap},i}(.05)$ at each $i\geq 3$. 
The averages of the extrapolated values are plotted in blue, and again, the yellow and green curves are obtained by adding or subtracting one standard deviation.

\textbf{Comments on results for IHS.} 
At a glance, Figure~\ref{fig:ihs} shows that the extrapolated estimate stays on track with the ideal benchmark, and is a nearly unbiased estimate of $\ve_{\text{IHS},i}(.05)$, for $i=3,\dots,10$. An interesting feature of the plots is how much the convergence rate of IHS depends on $m$.
Specifically, we see that after 10 iterations, the choice of $m=10d$ versus~$m=50d$ can lead to a difference in accuracy that is \emph{4 or 5 orders of magnitude}.
This sensitivity to $m$ illustrates why selecting $t$ is a non-trivial issue in practice, and why the extrapolated estimate can provide a valuable source of extra information.

\section{Conclusion}
We have proposed a systematic approach to answer a very practical question that arises for randomized LS algorithms: ``How accurate is a given solution?''
A distinctive aspect of the method is that it leverages the bootstrap --- a tool ordinarily used for statistical inference --- in order to serve a computational purpose. To our knowledge, it is also the first error estimation method for randomized LS that is supported theoretical guarantees. Furthermore, the method does not add much cost to an underlying sketching algorithm, and it has been shown to perform well on several examples.

\acks{%
Lopes is partially supported by NSF grant DMS-1613218.}

\appendix

\section*{Outline of Appendices}

The proof of Theorem~\ref{thm:main} is decomposed into two parts, with the bounds for IHS and CS being handled in Appendices~\ref{app:ihs} and~\ref{app:cs} respectively.

\section{Proof of Theorem~\ref{thm:main} for Iterative Hessian Sketch}\label{app:ihs}

To make the structure of the proof clearer, the main ingredients are combined in Appendix~\ref{sec:ihs_highlevel}. The lower-level arguments are given in Appendix~\ref{app:IHSlemmas}.

\paragraph{Remark on notation.} Since we will often need to condition on the sketching matrices $S_1,\dots,S_{t}$ in the IHS algorithm, we define $\mathcal{S}_k:=\{S_1,\dots,S_k\}$ for any $1\leq k\leq t$, and put $\mathcal{S}_{0}:=\emptyset$.

\subsection{High-level proof of the bound~\eqref{eqn:ihs_thm}}\label{sec:ihs_highlevel}
For any $\tau\in\R$, define the conditional distribution function
$$ F_n(\tau):=\P\Big(\|\hat x_t-x_{\opt})\|_{\circ} \leq \tau\Big\bracevert \S_{t-1}\Big).$$
%
%
%
Next, let $\ve_1^*,\dots,\ve_B^*$ be the samples generated by Algorithm \ref{alg:ihs}, and define the empirical distribution function
$$\hat F_{n,B}(\tau):=\ts\frac 1 B\tsum_{l=1}^B 1\{\ve_l^*\leq \tau\}.$$
%
In Proposition~\ref{prop:IHS} below, we show that  as $n\to\infty$,
\begin{equation}\label{eqn:ihsbootunif}
\sup_{\tau\in\R}\Big|\hat F_{n,B}(\tau)-F_n(\tau)\Big|\to 0 \text{ \ \ \ in \ $\P(\cdot\,|\S_{t-1})$-probability}.
\end{equation}
Establishing this limit is the most difficult part of the proof.
Next, for any number $p\in(0,1)$, and any distribution function $G$, define the quantile function $G^{-1}(p)=\inf\{\tau: G(\tau)\geq p\}$. 
Using this definition, as well as the limit~\eqref{eqn:ihsbootunif}, it follows that for any fixed $\delta\in(0,1-\alpha)$, the event
\begin{equation}\label{eqn:Edef}
\mathcal{E}_n(\delta):=\Big\{\hat \ve_t(\alpha) \geq F_n^{-1}(1-\alpha-\delta)\Big\}
\end{equation}
satisfies
\begin{equation}\label{eqn:Elim}
\P(\mathcal{E}_n(\delta)|\S_{t-1})\to 1,
\end{equation}  
and handful of details for checking this are given immediately after the end of this proof.
We now use the event $\mathcal{E}_n(\delta)$ to derive an upper bound on the probability \smash{$\P\big(\|\hat x_t-x_{\opt}\|_{\circ}>\hat \ve_t(\alpha)\big|\S_{t-1}\big)$.} From the definition of $\mathcal{E}_n(\delta)$, it follows that
\begin{small}
	\begin{align*}
	& \P\Big(\|\hat x_t-x_{\opt}\|_{\circ}>\hat \ve_t(\alpha)\Big|\S_{t-1}\Big)\\
	& =  \; \P\Big(\Big\{\|\hat x_t-x_{\opt}\|_{\circ}>\hat \ve_t(\alpha)\Big\}\cap \mathcal{E}_n(\delta)\Big|\S_{t-1}\Big) \, + \, \P\Big(\Big\{\|\hat x_t-x_{\opt}\|_{\circ}>\hat \ve_t(\alpha)\Big\}\cap \mathcal{E}_n^c(\delta)\Big|\S_{t-1}\Big)\\
	& \leq  \; \P\Big(\|\hat x_t-x_{\opt}\|_{\circ}>F_n^{-1}(1-\alpha-\delta)\Big|\S_{t-1}\Big) \, + \, \P(\mathcal{E}_n^c(\delta)|\S_{t-1})\\
	& = \; 1-  F_n\Big(F_n^{-1}(1-\alpha-\delta)\Big) \, + \, \P(\mathcal{E}_n^c(\delta)|\S_{t-1})\\
	& \leq \; \alpha+\delta+\P(\mathcal{E}_n^c(\delta)|\S_{t-1}),
	\end{align*}
\end{small}%
and in the last step we have used the basic fact $F_n(F_n^{-1}(p))\geq p$ for any $p\in(0,1)$. So, by taking the complement of the event $\{\|\hat x_t-x_{\opt}\|_{\circ}>\hat \ve_t(\alpha)\}$, the previous bounds give
\begin{equation}
\P\Big(\|\hat x_t-x_{\opt}\|_{\circ}\ \leq \ \hat \ve_t(\alpha)\Big|\S_{t-1}\Big) \geq 1-\alpha-\delta-\P(\mathcal{E}_n^c(\delta)|\S_{t-1}).
\end{equation}
Taking the expectation of both sides with respect to $\S_{t-1}$ leads to
\begin{equation}
\P\Big(\|\hat x_t-x_{\opt}\|_{\circ}\ \leq \ \hat \ve_t(\alpha)\Big) \geq 1-\alpha-\delta-\P(\mathcal{E}_n^c(\delta)).
\end{equation}
Then, taking $\liminf_{n\to\infty}$ on both sides with $\delta$ held fixed, and noting that that $\P(\mathcal{E}_n^c(\delta))\to 0$ (by the dominated convergence theorem and the limit~\eqref{eqn:Elim}), we obtain
\begin{equation}
\liminf_{n\to\infty} \P\Big(\|\hat x_t-x_{\opt}\|_{\circ}\ \leq \ \hat \ve_t(\alpha)\Big) \geq 1-\alpha-\delta.
\end{equation}
Since the left side above does not depend on the arbitrarily small number $\delta$, it follows that
\begin{equation}
\liminf_{n\to\infty} \P\Big(\|\hat x_t-x_{\opt}\|_{\circ}\ \leq \ \hat \ve_t(\alpha)\Big) \geq 1-\alpha,
\end{equation}
and this implies the inequality~\eqref{eqn:ihs_thm}.\qed

\paragraph{Details for showing~\eqref{eqn:Elim}.}
Due to the fact that $\hat\ve_t(\alpha)$ can be expressed as $\hat F_{n,B}^{-1}(1-\alpha)$, we have the basic inequality \smash{$\hat F_{n,B}(\hat\ve_t(\alpha))\geq 1-\alpha$}. Also, if we define $\hat \delta_n:=|\hat F_{n,B}(\hat\ve_t(\alpha))-F_n(\hat \ve_t(\alpha))|$, then
\begin{equation}\label{eqn:basicbound}
\begin{split}
F_n(\hat\ve_t(\alpha)) &\geq  \hat F_{n,B}(\hat\ve_t(\alpha))-\hat \delta_n\\[0.2cm]
&\geq 1-\alpha-\hat\delta_n.
\end{split}
\end{equation}
Next, the limit~\eqref{eqn:ihsbootunif} ensures that for any fixed $\delta\in(0,1-\alpha)$, the event $\mathcal{E}_n'(\delta):=\{\hat\delta_n\leq \delta\}$ satisfies $\P(\mathcal{E}_n'|\S_{t-1})\to 1$.
Furthermore, in light of the inquality~\eqref{eqn:basicbound} it is simple to check that $\mathcal{E}_n'(\delta)\subset\mathcal{E}_n(\delta)$, which implies \eqref{eqn:Elim}.\\

\begin{proposition}\label{prop:IHS}
	If the conditions of Theorem 1 hold, then the limit~\eqref{eqn:ihsbootunif} holds.
\end{proposition}

\proof Let $\|\hat x_t^*-\hat x_t\|_{\circ}$ be a bootstrap sample generated by Algorithm \ref{alg:ihs}, and for any $\tau\in\R$, define the conditional distribution function
\begin{equation}
\hat F_n(\tau):=\P\Big(\|\hat x_t^*-\hat x_t\|_{\circ}\leq \tau\Big|\S_{t}\Big).
\end{equation}
(Note that the set $\S_{t}$ has been conditioned on here, which means that $\hat F_n$ is a random function with respect to $\P(\cdot\, |\S_{t-1})$.)
Another important observation is that the bootstrap samples $\ve_1^*,\dots,\ve_B^*$ may be regarded as i.i.d.~draws from $\hat F_n$.
Due to the Dvoretzky-Kiefer-Wolfowitz inquality~\citep{Dvoretzky,Massart}, if $B\to\infty$ with $n\to\infty$, then
\begin{equation}
\sup_{\tau\in\R}\Big| \hat F_{n,B}(\tau)-\hat F_n(\tau)\Big|\to 0 \text{ \ \ \  in \ \  $\P(\cdot\,|\S_{t-1})$-probability.}
\end{equation}
(Note that this holds regardless of the rate at which $B$ diverges, and so no conditions on the relative sizes of $B$ and $n$ are needed.)
So, due to the simple inequality
$$
\sup_{\tau\in\R}\Big| \hat F_{n,B}(\tau)- F_n(\tau)\Big| \ \leq \ \sup_{\tau\in\R}\Big| \hat F_{n,B}(\tau)- \hat F_n(\tau)\Big| \ + \  \sup_{\tau\in\R}\Big|  \hat F_{n}(\tau)- F_n(\tau)\Big|,$$
the proof reduces to showing that
\begin{equation}
\sup_{\tau\in\R}\Big| \hat F_{n}(\tau)-F_n(\tau)\Big|\to 0 \text{ \ \ \  in \ \  $\P(\, \cdot \, |\S_{t-1})$-probability,}
\end{equation}
and this is the core aspect of the proof. This limit follows directly from Lemma~\ref{lem:mainunif}, which can be found at the end of the next subsection. (Prior to Lemma~\ref{lem:mainunif}, there are three other lemmas that assemble the main arguments.)\qed

\subsection{Lemmas supporting the proof of Proposition~\ref{prop:IHS} }\label{app:IHSlemmas}
In this section we will use some specialized notation. In addition, our proofs will rely on the convergence of conditional distributions, as reviewed below.

\paragraph{Notation for vectors and matrices.} We use $e_1,\dots,e_d$ to refer to the standard basis vectors in $\R^d$. Next, we define two basic operations $\uvec(\cdot)$ and $\sym(\cdot)$ on matrices and vectors. For a \emph{symmetric} matrix $M\in\R^{d\times d}$, let $\uvec(M)\in\R^{d(d+1)/2}$ be the vector obtained by extracting the upper triangular portion of $M$, where the entries of $\uvec(M)$ are ordered row-wise (starting from the first row). For example, 
$$M=\begin{pmatrix} m_{11} & m_{12} \\ m_{21} & m_{22} \end{pmatrix} \ \ \  \Longrightarrow \ \ \ \uvec(M)=(m_{11},m_{12},m_{22}).$$
Next, for  any vector $u\in\R^{d(d+1)/2}$, let $\sym(u)$ be the unique symmetric matrix in $\R^{d\times d}$ that satisfies $\uvec(\sym(u))=u$. For example,
$$u=(u_1,u_2,u_3) \ \ \ \Longrightarrow \ \ \ \text{sym}(u)=\begin{pmatrix} u_1 & u_2 \\ u_2 & u_3 \end{pmatrix}.$$
Define the normalized matrix $\bar A_n := \frac{1}{\sqrt n} A$, as well as the following analogues of $H_n=\frac 1n A\ttop A$,
\begin{equation}\label{eqn:Hdefs}
\tilde H_n:=\ts \frac 1n A\ttop S_t\ttop  S_t A \ \ \ \ \text{ and } \ \ \  \tilde H_n^* :=\ts\frac 1n A\ttop S_t^{*\top} S_t^*A.
\end{equation}
Lastly, when referring to the rows of $\sqrt m S_t$, we will omit the dependence on $t$ and write simply $s_1,\dots,s_m$ for ease of notation.
\paragraph{Convergence of conditional distributions.} If a sequence of random vectors $V_n$ converges in distribution to a random vector $V$, we write $\mathcal{L}(V_n)\xrightarrow{ \ d \ } \mathcal{L}(V)$.  In some situations, we will also need to discuss convergence of conditional distributions. To review the meaning of this notion, let $d_{\text{LP}}(\mathcal{L}(V_n),\mathcal{L}(V))$ denote the L\'evy-Prohorov distance~\citep[p.~394]{Dudley} between the distributions $\L(V_n)$ and $\L(V)$, and note the basic fact that $d_{\text{LP}}(\mathcal{L}(V_n),\mathcal{L}(V))\to 0$ if and only if $\mathcal{L}(V_n)\xrightarrow{ \ d \ } \mathcal{L}(V)$. Now, suppose $U_n$ is another sequence of random vectors, and let $d_{\text{LP}}(\mathcal{L}(V_n|U_n),\mathcal{L}(V|U_n))$ denote the $d_{\text{LP}}$ distance between $\mathcal{L}(V_n|U_n)$ and $\mathcal{L}(V|U_n)$, which are random probability distributions. Likewise, the sequence $\{d_{\text{LP}}(\mathcal{L}(V_n|U_n),\mathcal{L}(V|U_n))\}_{n=1}^{\infty}$ may be regarded as a sequence of scalar random variables, and if it happens that this sequence converges to 0 in probability, then we say `$\mathcal{L}(V_n|U_n)\xrightarrow{  \ d  \ } \mathcal{L}(V|U_n) \text{ in probability}$'.

The rest of this subsection consists of the four lemmas needed to prove Proposition~\ref{prop:IHS}. 


\begin{lemma}\label{lem:basicclt}
	Suppose the conditions of Theorem 1 hold. Then, there is a mean-zero random vector $V\in\R^{d(d+1)/2}$ with a multivariate normal distribution and a positive definite covariance matrix, such that as $n\to\infty$,
	\begin{equation}\label{eqn:clt}
	\mathcal{L}\big\{\sqrt m(\uvec(\tilde H_n)-\uvec(H_n))\big\}\xrightarrow{ \ d  \ }\mathcal{L}(V).
	\end{equation}
\end{lemma}
\noindent \emph{Proof.} Due to the Cram\'er-Wold theorem~\citep{Vaart}, it sufficient to show that for any fixed non-zero vector $c\in\R^{d(d+1)/2}$, the scalar random variable  $\langle \sqrt m(\uvec(\tilde H_n)-\uvec(H_n)), c\rangle$ converges in distribution to a zero-mean Gaussian random variable with positive variance. It is clear that for any such vector $c$, there is a corresponding upper triangular matrix $C\in\R^{d\times d}$ such that
\begin{equation}
\small
\begin{split}
\langle \sqrt m(\uvec(\tilde H_n)-\uvec(H_n)), c\rangle \ & = \ \sqrt m\big(\tr(\tilde H_n C)-\tr(H_nC)\big)\\[0.2cm]
& = \frac{1}{\sqrt m} \sum_{i=1}^m\Big( s_i\ttop \bar A_nC\bar A_n\ttop s_i - \tr(\bar A_nC\bar A_n\ttop)\Big),\\[0.2cm]
& = \frac{1}{\sqrt m} \sum_{i=1}^m\xi_{i,n}
\end{split}
\end{equation}
where we define the random variable $\xi_{i,n}:=s_i\ttop \bar A_n C\bar A_n\ttop s_i - \tr(\bar A_n C\bar A_n\ttop)$.
%
%
%
It is also clear that $\xi_{1,n},\dots,\xi_{m,n}$ are i.i.d.~with mean zero. 

As a preparatory step towards applying the central limit theorem, we now show that $\var(\xi_{1,n})$ converges to a positive limit. Because each vector $s_i$ is composed of i.i.d.~random variables, we may use an exact formula for the variance of quadratic forms~\citep[eqn.\,1.15]{Bai:Silverstein:2004}, which leads to
\begin{align}
\var(\xi_{1,n})&=\var\Big(s_1\ttop \bar A_nC\bar A_n\ttop s_1\Big)\label{eqn:hello}\\[0.2cm]
&= 2\big\|\bar A_nC\bar A_n\ttop \big\|_F^ 2+(\kappa-3)\sum_{j=1}^n (e_j\ttop \bar A_nC\bar A_n\ttop e_j)^2,\label{eqn:hello2}
\end{align}
where we recall that $\kappa:=\E[s_{i,j}^4]$ does not depend on $n$.
Next, observe that the relation $\|\bar A_nC\bar A_n\ttop \|_F^2=\tr(CH_nCH_n)$ implies that  $\|\bar A_nC\bar A_n\ttop\|_F^2\to \|H_{\infty}^{1/2}C H_{\infty}^{1/2}\|_F^2$, and also, the second term in line~\eqref{eqn:hello2} converges to a limit, say $(\kappa-3)\ell(C)$, due to Assumption 1. Hence,
\begin{equation}\label{eqn:varlimit} 
\var(\xi_{1,n}) \to 2\|H_{\infty}^{1/2}C H_{\infty}^{1/2}\|_F^2 +(\kappa-3)\ell(C).
\end{equation}

Now that we have shown $\var(\xi_{1,n})$ converges to a limit, we verify that this limit is positive. Since we assume $\kappa>1$, it is clear that $\kappa-3> \e_0-2$ for some fixed $\e_0\in(0,1)$. Also, since the second term in line~\eqref{eqn:hello2} represents the sum of the squares of the diagonal entries of $\bar A_n C \bar A_n\ttop$, the sum of the two terms must be at least $\e_0\|\bar A_n C\bar A_n\ttop\|_F^2$.  Therefore, the limit of $\var(\xi_{1,n})$ is lower-bounded by $\e_0\|H_{\infty}^{1/2}C H_{\infty}^{1/2}\|_F^2$, and because $H_{\infty}$ is positive definite, this lower bound is positive when $C\neq 0.$

Given that $\var(\xi_{1,n})$ converges to a positive limit, we now apply the central limit theorem. More specifically, since the common distribution of the variables $\{\xi_{1,n},\dots,\xi_{m,n}\}$ changes with $n$, we use the Lindeberg central limit theorem for triangular arrays~\citep[Prop. 2.27]{Vaart}. In addition to the existence of a limit for $\var(\xi_{1,n})$, this theorem requires that the limit $\E\big[ \xi_{1,n}^2 1\{|\xi_{1,n}| >\e\sqrt m\}\big]\to 0$
holds for any fixed $\e>0$ as $n\to\infty$. To verify this condition, the Cauchy-Schwarz inequality gives
\begin{equation}\label{eqn:lindebergbound}
\E\big[ \xi_{1,n}^2 1\{|\xi_{1,n}| >\e\sqrt m\}\big] \ \leq \sqrt{\E[\xi_{1,n}^4]}\cdot \sqrt{\P(\xi_{1,n}>\e\sqrt m)}.
\end{equation}
In turn, using a classical bound for moments of random quadratic forms~~\citep[Lemma B.26]{Bai:Silverstein:2010}, and the assumption that $\E[s_{i,j}^{8}]<\infty$, it is straightforward to check that $\E[\xi_{1,n}^4]=\mathcal{O}(1)$. Also, the condition $\P(\xi_{1,n}>\e\sqrt m)\to 0$ follows from Chebychev's inequality and the limit~\eqref{eqn:varlimit}. 
%
%
Therefore, the Lindeberg central limit theorem implies
\begin{equation}\label{eqn:lindeberg}
\mathcal{L}(\langle \sqrt m(\uvec(\tilde H_n)-\uvec(H_n)), c\rangle ) \xrightarrow{ \ d  \ } N(0,\sigma^2(C)),
\end{equation}
where we put $\sigma^2(C):=2\|H_{\infty}^{1/2}CH_{\infty}^{1/2}\|_F^2+(\kappa-3)\ell(C)$. This proves the limit~\eqref{eqn:clt}.\qed

\begin{lemma}\label{lem:bootclt}
	Suppose the conditions of Theorem 1 hold, and let $V$ be the random vector in statement of Lemma~\ref{lem:basicclt}. Then, as $n\to\infty$,
	\begin{equation}\label{eqn:bootclt}
	\mathcal{L}\big\{\sqrt m(\uvec(\tilde H_n^*)-\uvec(\tilde H_n))\big\bracevert S_t\big\}\xrightarrow{ \ d  \ }\mathcal{L}(V), \  \text{ in probability}.
	\end{equation}
\end{lemma}
\paragraph{Remark.} Note that the second limit holds in probability because  $\mathcal{L}\big\{\sqrt m(\uvec(\tilde H_n^*)-\uvec(\tilde H_n))\big\bracevert S_t\big\}$ is a random probability distribution that depends on $S_t$.

\proof
The overall approach is similar to the proof of Lemma~\ref{lem:basicclt}. If we let $s_1^*,\dots,s_m^*$ be drawn with replacement from $\{s_1,\dots,s_m\}$, then it is simple to check that $\tilde H_n$ can be represented as 
$$\tilde H_n^*=\frac 1m \sum_{i=1}^m \bar A_n\ttop s_i^*s_i^{*\top} \bar A_n.$$
Accordingly, for any $c\in\R^{d(d+1)/2}$ we have
\begin{equation}\label{eqn:bootalgebra}
\begin{split}
\Big\langle \sqrt m(\uvec(\tilde H_n^*)-\uvec(\tilde H_n)), c\Big\rangle \ & = \ \frac{1}{\sqrt m} \sum_{i=1}^m \xi_{i,n}^*,
\end{split}
\end{equation}
where $\xi_{i,n}^*:= s_i^{*\top}\bar A_n C\bar A_n\ttop s_i^* - \tr(C\tilde H_n)$, and $C$ is the upper-triangular matrix associated with $c$. Our goal is now to show that conditionally on the matrix $S_t$, the sum $\frac{1}{\sqrt m} \sum_{i=1}^m \xi_{i,n}^*$ satisfies the conditions of the Lindeberg central limit theorem (in probability), which will lead to the desired limit~\eqref{eqn:bootclt}. To do this, first observe that conditionally on $S_t$, the random variables $\xi_1^*,\dots,\xi_m^*$ are i.i.d., and satisfy
$\E[\xi_{i,n}^*|S_t] = 0$.
It remains to verify the following two conditions,
\begin{equation}\label{eqn:bootvarlimit}
\var(\xi_{1,n}^*|S_t)\to \sigma^2(C) \ \text{ in probability}, 
\end{equation}
where $\sigma^2(C)$ is as defined beneath line~\eqref{eqn:lindeberg}, and also
\begin{equation}\label{eqn:bootlindeberg}
\E[(\xi_{1,n}^*)^21\{|\xi_{1,n}^*>\e \sqrt m\} \, |S_t] \to 0 \text{ \ \ \ in probability},
\end{equation}
for any fixed $\e>0$.

To verify the limit~\eqref{eqn:bootvarlimit}, note that because $\xi_{1,n}^*,\dots,\xi_{m,n}^*$ can be viewed as samples with replacement from the set $\{\xi_{1,n},\dots,\xi_{m,n}\}$, it follows that
\begin{equation}
\var(\xi_{1,n}^*|S_t)=\hat\varsigma_n^{\, 2},
\end{equation}
where $\hat \varsigma_n^2$ denotes the sample variance
$\hat\varsigma_n^{\, 2}:=\frac 1m \sum_{i=1}^m (\xi_{i,n}-\bar\xi)^2,$
and $\bar \xi:=\frac 1m \sum_{i=1}^m \xi_{i,n}$.
It is clear that $\hat \varsigma^{\, 2}_n$ is asymptotically unbiased for $\sigma^2(C)$, since
$\E[\hat \varsigma^{\, 2}_n] =\ts\frac{m-1}{m} \var(\xi_{1,n})$.
So, to show that $\hat\varsigma^{\, 2}_n$ converges to $\sigma^2(C)$ in probability, it is enough to show that $\var(\hat\varsigma^{\, 2}_n)$ converges to 0. Using a classical formula for the variance of $\hat \varsigma_n$~\citep[p.164]{Kenney}, it is simple to obtain the bound
\begin{equation}
\var(\hat\varsigma^{\, 2}_n) \ = \mathcal{O}\Big( \frac{1}{n} \frac{\mu_{4,n}}{\var(\xi_{1,n})^2}\Big),
\end{equation}
where  $\mu_{4,n}$ is the fourth central moment of $\xi_{1,n}$, i.e.
$$\mu_{4,n}=\E\Big[ \big(\xi_{1,n}-\E[\xi_{1,n}]\big)^4\Big].$$
Using a general bound for the moments of quadratic forms~\citep[Lemma B.26]{Bai:Silverstein:2010}, this quantity can be bounded as
\begin{equation}
\mu_{4,n} =\mathcal{O}\Big(\tr(M_n)^2+\tr(M_n^2)\Big)
\end{equation}
where $M_n:=(\bar A_n C\bar A_n\ttop)^2$.
Since both of the traces above can be expressed in terms of the matrix $C\bar A_n\ttop \bar A_n$, which converges to $C H_{\infty}$, it follows that $\mu_{4,n}=\mathcal{O}(1)$. Also, it was shown in the proof of Lemma~\ref{lem:basicclt} that $\var(\xi_{1,n})$ has a positive limit. Altogether, this completes the work needed to prove the limit~\eqref{eqn:bootvarlimit}.

Finally, to verify limit~\eqref{eqn:bootlindeberg}, observe that since  $\E\big[(\xi_{1,n}^*)^21\{|\xi_{1,n}^*>\e \sqrt m\} \, \big|S_t\big]  $ is a non-negative random variable, Markov's inequality ensures that convergence to 0 in expectation implies convergence to 0 in probability. Using the fact that $\xi_{1,n}^*$ is sampled with replacement from the set $\{\xi_{1,n},\dots,\x_{m,n}\}$, we have
\begin{equation}
\E\big[(\xi_{1,n}^*)^21\{|\xi_{1,n}^*>\e \sqrt m\} \, \big|S_t\big]  = \frac 1m \sum_{i=1}^m \xi_{i,n}^2 1\{\xi_{i,n}>\e\sqrt m\},
\end{equation}
and so
$$\E\Big[ \E\big[(\xi_{1,n}^*)^21\{|\xi_{1,n}^*>\e \sqrt m\} \, \big|S_t\big]\Big] = \E\big[\xi_{1,n}^21\{\xi_{1,n}>\e\sqrt m\}\big].$$
Consequently, the argument based on the bound~\eqref{eqn:lindebergbound} in the proof of Lemma~\ref{lem:basicclt} may be re-used to show that the right hand side above tends to 0.\qed

\paragraph{Remarks on notation.} For the following lemma, let $\mathcal{U}\subset \R^{d(d+1)/2}$ denote the set of all vectors $u$ that can be represented as $u=\uvec(M)$ for some symmetric invertible matrix $M\in\R^{d\times d}$.  In this notation, we define the map $\phi :\mathcal{U}\to\mathcal{U}$ by
$$\phi(u)=\uvec(\sym(u)^{-1}).$$

\begin{lemma}\label{lem:delta}
	Suppose the conditions of Theorem 1 hold. Then there is a mean-zero multivariate normal random vector $W\in\R^{d(d+1)/2}$ with a positive definite covariance matrix such that as $n\to\infty$,
	\begin{equation}\label{eqn:delta1}
	\L\Big\{\sqrt m \Big(\phi(\uvec(\tilde H_n))-\phi(\uvec(H_n))\Big)\Big\} \xrightarrow{ \ d  \ } \L(W),
	\end{equation}
	and
	\begin{equation}\label{eqn:delta2}
	\L\Big\{\sqrt m \Big(\phi(\uvec(\tilde H_n^*))-\phi(\uvec(\tilde H_n))\Big)\Big\bracevert S_t\Big\} \xrightarrow{ \ d  \ } \L(W) \  \text{ in probability}.
	\end{equation}
\end{lemma}

\proof Recall that we assume $\sqrt m(H_n-H_{\infty})\to 0$, where $H_{\infty}$ is positive definite. Since the map $\phi$ is differentiable, it follows from the delta method~\citep[Theorem 3.1]{Vaart} and Lemma~\ref{lem:basicclt} that
$$\mathcal{L}\big(\sqrt{m}(\phi(\uvec(\tilde H_n))-\phi(\uvec(H_n)))|S_t)\xrightarrow{ \ d  \ }\mathcal{L}(W), \text{ \ \ in probability} $$
where we define
$W:=\phi'_0(V),$
with  $V$ being the random vector in Lemma~\ref{lem:basicclt}, and $\phi'_0$  denoting the differential of $\phi$ at the point $\uvec(H_{\infty})$. 

Due to Lemma~\ref{lem:basicclt}, we know that $V$ has a multivariate normal distribution with mean zero and a positive-definite covariance matrix. Also, because the map $\phi$ and its inverse $\phi^{-1}$ are differentiable on $\mathcal{U}$, it follows that the differential $\phi'_0$ must be an invertible linear map on $\R^{d(d+1)/2}$. Consequently, the random vector $W=\phi_0'(V)$ has a positive definite covariance matrix.
Finally, the same reasoning can be used to obtain the limit~\eqref{eqn:delta2}, since the limit~\eqref{eqn:bootclt} holds almost surely along subsequences, and the delta method may be applied again with the map $\phi$ (cf.~\citet[Theorem 23.5]{Vaart}) .\qed

\begin{lemma}\label{lem:mainunif}
	Suppose the conditions of Theorem 1 hold, and let $Z=\sym(W)$, with $W$ being  random vector in the statement of Lemma~\ref{lem:delta}. Then, for almost every sequence of sets $\S_{t-1}$, the following limit holds as $n\to\infty$, 
	\begin{small}
	\begin{equation}\label{eqn:uniflim1}
	\sup_{\tau\in\R}\bigg|\P\Big( \sqrt m\|\hat x_t - x_{\textup{opt}}\|_{\circ} \leq \tau\big\bracevert \mathcal{S}_{t-1}\Big)-\P\Big(\ts\frac{1}{n}\|Z g_{t-1}\|_{\circ}\leq \tau \big\bracevert \mathcal{S}_{t-1}\Big)\bigg|\to 0.
	\end{equation}
	\end{small}%
	Furthermore,
	\begin{small}
	\begin{equation}\label{eqn:uniflim2}
	\sup_{\tau\in\R}\bigg|\P\Big( \sqrt m\|\hat x_t^* - \hat x_t\|_{\circ} \leq \tau\big\bracevert \mathcal{S}_t \Big)-\P\Big(\ts\frac{1}{n}\|Z g_{t-1}\|_{\circ}\leq \tau\big\bracevert \mathcal{S}_{t-1}\Big)\bigg|\to 0, \ \text{ in \  $\P(\cdot\,|\S_{t-1})$-probability}.
	\end{equation}
\end{small}%
\end{lemma}

\proof  We first prove the limit~\eqref{eqn:uniflim1}. 
For any fixed vector $v\in\R^d$, and fixed scalar $\tau$, define the set $\mathcal{C}(v,\tau)\subset\R^{d(d+1)/2}$ to contain the vectors $u$ satisfying $\|\sym(u)v\|_{\circ}\leq \tau$. 
Based on this definition of $\mathcal{C}(v,\tau)$, the following events are equal
\begin{equation}
\big\{\ts\frac{1}{n}\|Z g_{t-1}\|_{\circ}\leq \tau\big\} = \big\{\uvec(Z)\in \mathcal{C}(g_{t-1},n\tau)\big\}.
\end{equation}
Next, using the relation
$$\hat x_t-x_{\opt} = -\ts\frac{1}{n}\big(\tilde H_n^{-1}-H_n^{-1}\big) g_{t-1},$$
it straightforward to check  that the following events are also equal
\begin{equation}\label{eqn:equiv1}
\Big\{\sqrt m\|\hat x_t-x_{\opt}\|_{\circ}\leq \tau\Big\} = \Big\{\sqrt m\Big(\phi(\uvec(\tilde H_n))-\phi(\uvec(H_n))\Big)\in\mathcal{C}(g_{t-1},n\tau)\Big\}.
\end{equation}
To proceed, we make use of the observation that the set $\mathcal{C}(g_{t-1},n \tau)$ is always convex and (Borel) measurable. Likewise, if we let $\mathscr{C}_{\text{convex}}$ denote the collection of all measurable convex subsets of $\R^{d(d+1)/2}$, it follows that the following supremum over $\tau\in\R$

\begin{equation}
\sup_{\tau\in\R}\bigg| \P\Big(\sqrt m\|\hat x_t-x_{\opt}\|_{\circ}\leq \tau\Big\bracevert \mathcal{S}_{t-1}\Big)-\P\Big(\ts\frac{1}{n}\|Z g_{t-1}\|_{\circ}\leq \tau\Big\bracevert \mathcal{S}_{t-1}\Big)\bigg| 
\end{equation}
is upper bounded by the following supremum over $\mathcal{C}\in\mathscr{C}_{\text{convex}}$,
\begin{equation}\label{eqn:bigsup}
\ \sup_{\mathcal{C}\in\mathscr{C}_{\text{convex}}}\bigg| \P\Big(\sqrt m \Big(\phi(\uvec(\tilde H_n))-\phi(\uvec(H_n))\Big)\in\mathcal{C}\Big\bracevert \mathcal{S}_{t-1}\Big) -\P(\uvec(Z)\in\mathcal{C}\big\bracevert \mathcal{S}_{t-1})\bigg|.
\end{equation}
To conclude the proof of~\eqref{eqn:uniflim1}, it suffices to show that the previous expression converges to 0 as $n\to\infty$. For this purpose, we apply the general fact that if a sequence of random vectors $\zeta_n$ converges in distribution to a random vector $\zeta$, and if $\zeta$ has a multivariate normal distribution with a positive definite covariance matrix, then 
\begin{equation}\label{eqn:raolimit}
\ \sup_{\mathcal{C}\in\mathscr{C}_{\text{convex}}}\Big| \P(\zeta_n\in\mathcal{C})-\P(\zeta\in\mathcal{C})\Big| \to 0.
\end{equation}
(We refer to the book~\citet[Theorem 1.11]{Rao} for further details.)
Now, observe that $\tilde H_n$ is independent of $\mathcal{S}_{t-1}$, and Lemma~\ref{lem:delta} ensures that $\sqrt m \big(\phi(\uvec(\tilde H_n))-\phi(\uvec(H_n))\big)$ converges in distribution to $W=\uvec(Z)$, which is multivariate normal with a positive definite covariance matrix. Consequently, the conditioning on $\mathcal{S}_{t-1}$ may be dropped, and the limit~\eqref{eqn:raolimit} implies that the supremum in line~\eqref{eqn:bigsup} must converge to 0 as $n\to\infty$. Finally, the bootstrap limit~\eqref{eqn:uniflim2} may be proven by repeating the same argument in conjunction with~\eqref{eqn:delta2}.
\qed

\section{Proof of Theorem~\ref{thm:main} for Classic Sketch}\label{app:cs}

\subsection{High-level proof of the bound~\eqref{eqn:cs_thm}}\label{sec:cs_highlevel}
In analogy with Appendix~\ref{sec:ihs_highlevel}, let $\tau\in\R$, and define the following conditional distribution function
$$ G_n(\tau):=\P\Big(\|\tilde x_t-x_{\opt}\|_{\circ} \leq \tau \Big).$$
Also, letting  $\ve_1^*,\dots,\ve_B^*$ denote the samples generated by Algorithm \ref{alg:cs}, define 
\begin{align}
\hat G_n(\tau) &:=\P\Big(\|\tilde x^*-\tilde x\|_{\circ}\leq \tau\Big\bracevert  S\Big),\\[0.3cm]
\hat G_{n,B}(\tau) &:=\ts\frac 1 B\tsum_{l=1}^B 1\{\ve_l^*\leq \tau\}.
\end{align}
By using these functions in place of their IHS counterparts $F_n(\tau)$, $\hat F_n(\tau)$, and $\hat F_{n,B}(\tau)$, the argument at the beginning of Appendix~\ref{sec:ihs_highlevel} can be essentially repeated to reach the conclusion
\begin{equation}
\liminf_{n\to\infty} \P\Big(\|\tilde x_t-x_{\opt}\|_{\circ}\ \leq \ \tilde \ve(\alpha)\Big) \geq 1-\alpha,
\end{equation}
which implies the desired inequality~\eqref{eqn:cs_thm}. The only part of the argument that needs to be updated is to prove the analogue of Proposition~\ref{prop:IHS} for the case of CS. In other words, it suffices to show that
\begin{equation}\label{eqn:csbootunif}
\sup_{\tau\in\R}\Big|\hat G_{n}(\tau)-G_n(\tau)\Big|\to 0 \text{ \ \ \ in probability.}
\end{equation}
Proving this limit will be handled with Proposition~\ref{prop:csconc} below.\qed

\paragraph{Remarks on notation.} The proof of Proposition~\ref{prop:csconc} relies on the following preliminary result. To introduce some notation, we will use the normalized gradient vector $\texttt{g}_n:=\ts\frac 1n A\ttop b$, and the analogues
\begin{equation*}
\tilde{\texttt{g}}_n := \ts\frac 1n \tilde A\ttop \tilde b, \text{ \ \ \ and \ \ \ }
\tilde{\texttt{g}}_n ^*:=\ts\frac 1n (\tilde A^*)\ttop (\tilde b^*).
\end{equation*}
Note that $\texttt{g}_n$ is not the same as the gradient $g_t$ used previously in the context of IHS. One additional detail to clarify is that in this section, we will overload the notation introduced in line~\eqref{eqn:Hdefs}. Specifically, we re-define $\tilde H_n$ and $\tilde H_n^*$ in terms of the single sketching matrix $S$ for CS and its resampled version $S^*$ CS (rather than the matrices $S_t$ and $S_t^*$ used in the context of IHS). That is,
\begin{equation}
\tilde H_n:=\ts \frac 1n A\ttop S\ttop  S A \ \ \ \ \text{ and } \ \ \  \tilde H_n^* :=\ts\frac 1n A\ttop S^{*\top} S^*A.
\end{equation}
Furthermore, the matrix $S$ has the same distribution as $S_t$, and so the Lemmas~\ref{lem:basicclt},~\ref{lem:bootclt} and~\ref{lem:delta}, involving $\tilde H_n$ and $\tilde H_n^*$, apply to the CS context with no changes.

\begin{lemma}\label{lem:fullsketchclt}
	Suppose the conditions of Theorem 1 hold. Then, there is a mean-zero random vector $Y\in\R^d$ having a multivariate normal distribution and a non-zero covariance matrix, such that as $n\to\infty$,
	\begin{equation}
	\mathcal{L}(\sqrt m(\tilde x-x_{\opt})) \xrightarrow{ \ d   \ \ } \mathcal{L}(Y),
	\end{equation}
	and
	\begin{equation}\label{eqn:fullbootclt}
	\mathcal{L}(\sqrt m(\tilde x^*-\tilde x)|S)\xrightarrow{ \ d  \ }\mathcal{L}(Y), \  \text{ in probability}.
	\end{equation}
\end{lemma}

\proof
The proof of Lemmas~\ref{lem:basicclt} and~\ref{lem:bootclt} can be adapted to show that the following joint limits hold
\begin{equation}\label{eqn:vecclt}
\mathcal{L}\Big\{\sqrt m\Big(\big(\uvec(\tilde H_n),\tilde{\texttt{g}}_n \big)-\big(\uvec(H_n),\texttt{g}_n\big)\Big)\Big\}\xrightarrow{ \ d  \ } \mathcal{L}(V,U),
\end{equation}
and
\begin{equation}\label{eqn:vecbootclt}
\mathcal{L}\Big\{\sqrt m\Big(\big(\uvec(\tilde H_n^*),\tilde{\texttt{g}}_n ^*\big)-\big(\uvec(\tilde H_n),\tilde{\texttt{g}}_n\big) \Big)\Big\bracevert S\Big\}\xrightarrow{ \ d  \ } \mathcal{L}(V,U),
\end{equation}
where $U\in\R^d$ is a random vector such that the concatenated vector $(V,U)\in\R^{d(d+1)/2}\times \R^d$ has a mean-zero multivariate normal distribution with a non-zero covariance matrix. To proceed, recall that $\mathcal{U}\subset\R^{d(d+1)/2}$ denotes the set of vectors  that can be written as $\uvec(M)$ for some symmetric invertible matrix $M$. Also, recall that for any $u\in\mathcal{U}$, the expression $\sym(u)$ refers to the unique symmetric matrix in $\R^{d\times d}$ that satisfies $\uvec(\sym(u))=u$. Next, consider the map $\Phi:\mathcal{U}\times \R^d \to \R^d$ defined by
$$\Phi(u,v) = (\sym(u))^{-1}v,$$
as well as the following relations, which are straightforward to verify
\begin{equation}
\tilde x-x_{\opt} = \tilde H_n^{-1}\tilde{\texttt{g}}_n -H_n^{-1} \texttt{g}_n
\end{equation}
\begin{equation}
\ \ \ \ \ \  \ \ \ \tilde x^*-\tilde x = (\tilde H_n^*)^{-1}\tilde{\texttt{g}}_n ^* - \tilde H_n^{-1}\tilde{\texttt{g}}_n.
\end{equation}
These relations can be written in terms of the map $\Phi$ as
$$\sqrt m(\tilde x - x_{\opt}) = \sqrt m\Big(\Phi\big(\uvec(\tilde H_n), \tilde{\texttt{g}}_n \big) - \Phi\big(\uvec(H_n), \texttt{g}_n\big)\Big),$$
and
$$\sqrt m(\tilde x^* - \tilde x) = \sqrt m\Big(\Phi\big(\uvec(\tilde H_n^*), \tilde{\texttt{g}}_n ^*\big) - \Phi\big(\uvec(\tilde H_n), \tilde{\texttt{g}}_n \big)\Big).$$
Next, recall the assumptions $\sqrt m(H_n-H_{\infty})\to 0$ and $\sqrt m(\texttt{g}_n-\texttt{g}_{\infty})\to 0$, and note that the map $\Phi$ is differentiable. Consequently, it follows from the delta method~\citep[Theorem 3.1]{Vaart}, as well as the limit~\eqref{eqn:vecclt} that
$$\mathcal{L}(\sqrt m(\tilde x-x_{\opt})) \xrightarrow{ \  \ d  \ \ } \mathcal{L}(\Phi_{0}'(V,U)),$$
where $\Phi_0'$ denotes the differential of $\Phi$ evaluated at the point $(\uvec(H_{\infty}),\texttt{g}_{\infty})$. Furthermore, since $\Phi$ and $\Phi^{-1}$ are differentiable, it follows that $\Phi_0'$ is an invertible linear map, which implies that the random vector $\Phi_0'(V,U)$ has a non-zero covariance matrix (since $(V,U)$ does). Similarly, the delta method can be applied to the limit~\eqref{eqn:vecbootclt} to obtain
$$\mathcal{L}(\sqrt m(\tilde x^*-\tilde x)|S) \xrightarrow{ \ \ d  \ \ } \mathcal{L}(\Phi_{0}'(V,U)), \text{ in  probability}.$$
Finally, letting $Y=\Phi_0'(V,U)$ completes the proof.\qed

\begin{proposition}\label{prop:csconc}
	If the conditions of Theorem 1 hold, then the limit~\eqref{eqn:csbootunif} holds
\end{proposition}
\proof Let $Y$ be the random vector in the statement of Lemma~\ref{lem:fullsketchclt}. Combining that lemma with the continuous mapping theorem~\citep[Theorem 2.3]{Vaart}, and the fact that any norm $\|\cdot\|_{\circ}$ on $\R^d$ is continuous, we have
\begin{equation}
\mathcal{L}(\sqrt m\|\tilde x-x_{\opt}\|_{\circ}) \xrightarrow{ \ d   \ \ } \mathcal{L}(\|Y\|_{\circ}),
\end{equation}
and
\begin{equation}
\mathcal{L}\big(\sqrt m\|\tilde x^*-\tilde x\|_{\circ}\big\bracevert S\big)\xrightarrow{ \ d  \ }\mathcal{L}(\|Y\|_{\circ}), \  \text{ in probability}.
\end{equation}
Since the random vector $Y$ has a multivariate normal distribution with a non-zero covariance matrix, it is straightforward to show that the random variable $\|Y\|_{\circ}$ has a continuous distribution function.  So, it follows from Polya's theorem~\citep[Theorem B.7.7]{Bickel} that
\begin{equation}
\sup_{\tau\in\R}\Big|\P\Big(\sqrt m \|\tilde x-x_{\opt}\|_{\circ}\leq \tau\Big)- \P\big(\|Y\|_{\circ}\leq \tau\big)\Big| \to 0,
\end{equation}
and
\begin{equation}
\ \ \ \  \  \ \ \ \ \ \ \ \ \ \ \ \ \ \ \ \ \sup_{\tau\in\R}\Big|\P\Big(\sqrt m \|\tilde x^*-\tilde x\|_{\circ}\leq \tau\Big\bracevert S\Big)- \P\big(\|Y\|_{\circ}\leq \tau\big)\Big| \to 0 \text{ \ \  in probability},
\end{equation}
which implies the limit~\eqref{eqn:csbootunif} by the triangle inequality.\qed


\bibliography{bibs/matrix,bibs/bootstrap,bibs/bootLSbib}

\end{document}